%% file: main.tex
\documentclass{commonstack}

\usepackage[utf8]{inputenc}
\usepackage[T1]{fontenc}
\usepackage{array}
\usepackage{makecell}
\usepackage{textgreek}
\usepackage{subcaption}
\usepackage{amsmath}
\usepackage{amssymb}
\usepackage{mathtools}
\usepackage{amsthm}
\usepackage{colortbl}
\usepackage{wrapfig}
\usepackage{mathpazo}
\usepackage{algorithm}
\usepackage{algpseudocode}
\usepackage{listings}
\usepackage{multirow}
\usepackage{siunitx}
\usepackage{tikz}
\usetikzlibrary{arrows.meta,fit,positioning,backgrounds}
\usepackage{enumitem}
\usepackage{pifont}
\usepackage{tabularx}
\usepackage{adjustbox}
\usepackage{booktabs}
\usepackage{xurl}

\title{MERA: Model Evolution and Routing with Skill Adaptation for Agentic Systems at Scale}

\author{%
Yuhang Yao\textsuperscript{4,\ensuremath{\dagger}} \quad
Zeyu Wang\textsuperscript{6} \quad
Wanyi Chen\textsuperscript{2} \quad
Tongyun Yang\textsuperscript{3} \quad
Yuhang Han\textsuperscript{5} \\
Jie Xiao\textsuperscript{1} \quad
Chengke Bao\textsuperscript{1} \quad
Tianyi Zhao\textsuperscript{1} \quad
Lynn Ai\textsuperscript{1} \quad
Eric Yang\textsuperscript{1} \quad
Tianyu Shi\textsuperscript{1,\ensuremath{\dagger}}%
}
\affiliation{%
\textsuperscript{1}Gradient \quad
\textsuperscript{2}Soochow University \quad
\textsuperscript{3}Independent Researcher \quad
\textsuperscript{4}Carnegie Mellon University \quad \\
\textsuperscript{5}Shanghai Jiao Tong University \quad
\textsuperscript{6}University of California, Los Angeles \\
\textsuperscript{\ensuremath{\dagger}}Corresponding author%
}

\correspondence{yuhangyao8@gmail.com, tianyu@gradient.network}
\sourcecode{https://github.com/yh-yao/MERA-Evolve}

\abstract{\input{0_abstract}}

\begin{document}

\maketitle

\input{1_introduction}
\input{2_related_work}
\input{3_method}
\input{4_experiment}
\input{5_conclusion}

\clearpage
\bibliography{ref}

\clearpage
\input{appendix}

\end{document}

%% file: 0_abstract.tex
LLM agents execute heterogeneous sequences of model calls within a single task: some invocations require careful reasoning, while others are structured steps such as formatting or tool-argument construction. Prior routing methods exploit this asymmetry by assigning easy invocations to a cheaper small model and difficult ones to a large model. Such policies reduce inference cost, but they leave the small model's capability unchanged, so attainable savings remain bounded by the work the student can already solve. MERA instead improves the small model itself, using a single model invocation as the unit of adaptation. In each cycle, MERA replays failed student invocations to obtain execution-verified teacher demonstrations, distills recurring procedures into an iteratively updated SkillBook, and fine-tunes a student LoRA adapter via supervised learning and optional GRPO. Routing serves as supporting machinery for deployment: the improved student is served behind a cost-calibrated router with verifier-backed fallback, and a candidate SkillBook, adapter, or router is admitted only when joint replay preserves task quality. Empirically, four-cycle adaptation raises Qwen2.5-Coder-1.5B from 28.7\% to 49.7\% pass on held-out HumanEval$+$MBPP. Under verifier-backed fallback, the deployed policy retains 88.3\% pass at 60.8\% of always-Luna cost. On TAU-2, a fine-tuned Qwen3.5-2B improves from 14/35 to 18/35 and matches an unadapted 4B model. These results indicate that verifier-backed multi-cycle adaptation can increase small-model capability, rather than only routing around a fixed student.

%% file: 1_introduction.tex
\section{Introduction}
Language-model agents execute heterogeneous chains of model calls. A single task may contain hard reasoning or policy-sensitive decisions, but it also contains many structured steps: extraction, formatting, tool-argument construction, post-processing, clarification, and templated summaries. Running every invocation on the strongest model is reliable but expensive. Conversely, routing only once at the user-task level is too coarse, because the easy and hard parts of a workflow often appear side by side. Recent work has improved agents through prompting, tool use, planning, search, and agent optimization \citep{react2023,toolformer2023,lats2024,toolllm2023,agentoptimizer2024,gptswarm2024}, but production systems also need mechanisms for adapting the models, routes, and reusable skills inside the workflow.

This creates a practical tension for deployed agent systems. The traces needed for adaptation are naturally produced online, but directly changing the serving policy from raw traces is risky: a cheaper model may fail silently, a learned adapter may regress on rare cases, and a reusable template may only be safe under a narrow prompt signature. A useful evolution loop therefore needs to separate observation from admission. It should collect evidence at the granularity of individual invocations, update several candidate components, and admit the resulting runtime state only when replay shows that the combined policy still satisfies verification constraints.

MERA addresses this problem by treating a single model invocation as the unit of adaptation. At runtime, an input-only router chooses among strong, cheap, and specialized models; a skill layer can dispatch stable templates for recurring local structure; and a verifier protects quality through fallback. This design keeps serving simple: the router does not depend on hidden agent state, and unsafe down-routing is corrected by verification. The heavier logic is moved offline, where traces are replayed to decide which prompts are easy, which failures should become training examples, and which recurrent patterns are stable enough to become skills.

\begin{figure*}[t]
    \centering
    \resizebox{0.96\textwidth}{!}{\input{intro_overview_tikz}}
    \caption{Overview of MERA. Online traces drive scheduled SkillBook, LLM-update, and router tracks; their combined state is admitted through joint replay evaluation.}
    \label{fig:intro_overview}
\end{figure*}
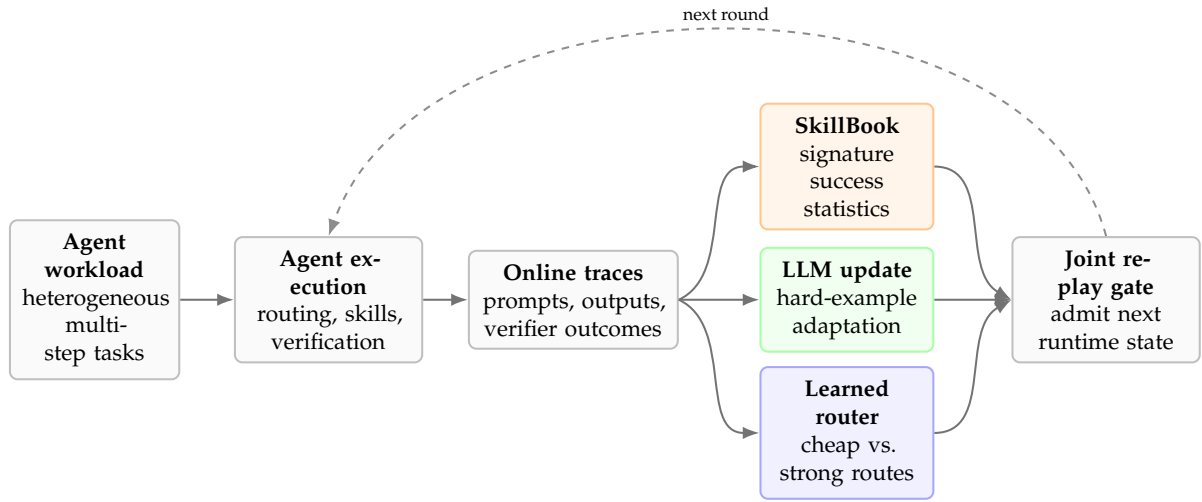

Figure~\ref{fig:intro_overview} shows the resulting feedback loop. Online traces drive update rounds in which SkillBook statistics, router training, and LLM adaptation share evidence. Within a cycle we use the dependency-respecting Skill $\rightarrow$ LLM $\rightarrow$ Router schedule: the adapter is trained with the current SkillBook procedure, and the router is fit last so its labels reflect current-cycle student outcomes. The combined state is admitted only after joint replay evaluation, so routing, skill promotion, and adapter updates are judged by their end-to-end effect rather than by isolated component metrics.

Our contribution is a verifier-backed multi-cycle protocol for improving the small model from shared executable traces; routing and SkillBook are supporting machinery for safe deployment. On 582 held-out HumanEval$+$MBPP tasks (3 seeds), four-cycle adaptation raises Qwen2.5-Coder-1.5B from 28.7\% to 44.2\% under SFT and to 49.7\% with matched SFT+GRPO. With executable verification and GPT-5.6 Luna fallback, the deployed policy retains 88.3\% pass at 60.8\% of always-Luna cost. On TAU-2, an adapted Qwen3.5-2B improves from 14/35 to 18/35 and matches an unadapted 4B endpoint, though the comparison is underpowered. A finance break-even analysis further connects adaptation cost to serving savings. MERA is therefore an evolution protocol first: it raises small-model capability rather than only routing around a fixed student.

%% file: intro_overview_tikz.tex
\begin{tikzpicture}[
  x=1cm, y=1cm,
  >=Latex,
  font=\small,
  block/.style={draw=black!25, rounded corners=3pt, thick, align=center, inner sep=5pt, minimum height=0.95cm},
  blueblock/.style={block, fill=blue!6, draw=blue!35},
  greenblock/.style={block, fill=green!6, draw=green!35},
  orangeblock/.style={block, fill=orange!8, draw=orange!45},
  grayblock/.style={block, fill=black!2},
  flow/.style={->, thick, draw=black!55},
  update/.style={->, thick, dashed, draw=black!45}
]

\node[grayblock, text width=2.0cm] (workload) at (0,0) {\textbf{Agent workload}\\heterogeneous\\multi-step tasks};
\node[grayblock, text width=2.25cm] (runtime) at (3.25,0) {\textbf{Agent execution}\\routing, skills,\\verification};
\node[grayblock, text width=2.55cm] (trace) at (6.65,0) {\textbf{Online traces}\\prompts, outputs,\\verifier outcomes};

\node[orangeblock, text width=2.05cm] (skills) at (10.45,1.85) {\textbf{SkillBook}\\signature success\\statistics};
\node[greenblock, text width=2.05cm] (models) at (10.45,0) {\textbf{LLM update}\\hard-example\\adaptation};
\node[blueblock, text width=2.05cm] (routing) at (10.45,-1.85) {\textbf{Learned router}\\cheap vs.\\strong routes};
\node[grayblock, text width=2.25cm] (replay) at (14.05,0) {\textbf{Joint replay gate}\\admit next\\runtime state};

\draw[flow] (workload) -- (runtime);
\draw[flow] (runtime) -- (trace);
\draw[flow] (trace.east) to[out=25,in=180] (skills.west);
\draw[flow] (trace.east) -- (models.west);
\draw[flow] (trace.east) to[out=-25,in=180] (routing.west);
\draw[flow] (skills.east) to[out=0,in=150] (replay.west);
\draw[flow] (models.east) -- (replay.west);
\draw[flow] (routing.east) to[out=0,in=-150] (replay.west);

\draw[update] (replay.north) to[out=105,in=70,looseness=0.95] node[above, font=\scriptsize] {next round} (runtime.north);
\end{tikzpicture}

%% file: 2_related_work.tex
\section{Related Work}

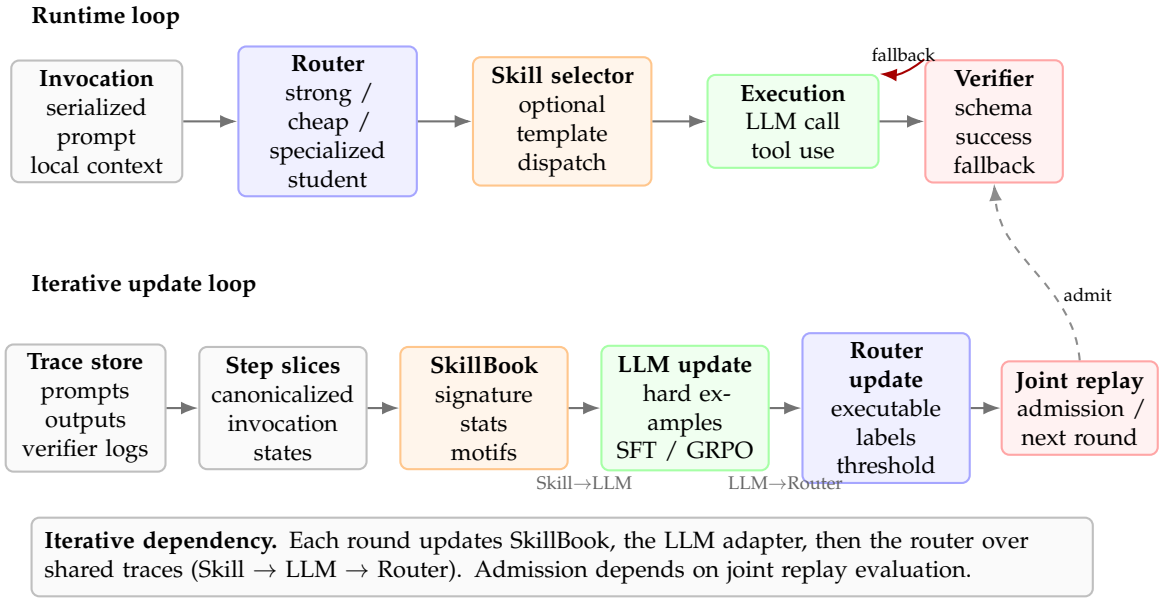
\begin{figure*}[t]
    \centering
    \resizebox{0.93\textwidth}{!}{\input{method_pipeline_tikz}}
    \caption{Detailed method view of MERA. Runtime routing remains input-only and verifier-protected; update tracks share traces and are admitted through joint replay.}
    \label{fig:system_diagram}
\end{figure*}

\textbf{LLM routing.}
Prior work on model routing studies how to dispatch requests across model pools under cost-quality trade-offs. FrugalGPT-style cascades, learned routers, and preference- or uncertainty-based selectors use prompt features, predicted quality, or explicit routing objectives to decide when a cheaper model is sufficient and when a stronger model is needed \citep{frugalgpt2024,routellm2024,dekoninck2025routing,bestroute2025}. Recent systems further consider upgraded model tiers, multi-round routing, and answer aggregation rather than one-shot dispatch \citep{llmat2025,routerr12025}. This literature establishes routing as a practical mechanism for reducing inference cost, but most settings route whole user requests or single-turn examples. The unit of decision is usually the prompt submitted to a model, not an internal invocation within a longer agent execution.

\textbf{Agent optimization.}
Agent research has focused on making agents more capable through tool use, reasoning-action loops, execution-time search, and automated optimization of agent graphs or functions \citep{react2023,toolformer2023,lats2024,toolllm2023,agentoptimizer2024,gptswarm2024}. Recent post-training and environment-based approaches also train agents from interaction traces or simulated feedback \citep{autopdl2025,agentgymrl2026}. These methods improve the policy, planner, or tool-use behavior of an agent. They typically treat the model choice, router, and reusable procedures as fixed engineering choices, or evaluate a single improved agent policy. They do not directly address how a deployed agent should use its own execution traces to update the model mixture and routing policy over time.

\textbf{Model specialization and reusable skills.}
Distillation, step-by-step supervision, small-model adaptation, reflective self-training, and explicit skill-library construction all aim to reuse behavior learned from stronger models or successful trajectories \citep{distillingstepbystep2023,agentr2025,skillx2026}. These methods show that smaller models and externalized skills can capture useful structure. However, their success is often reported as a standalone training or benchmark result. In deployed agent systems, specialization is conditional: a student or skill is useful only on slices where it remains reliable, and failures may need to trigger fallback rather than silently enter production.

\textbf{Evaluation for tool-using agents.}
Benchmarks for grounded web interaction, tool use, multi-hop tools, agent-user interaction, computer control, and software engineering provide increasingly realistic environments for measuring agent behavior \citep{webshop2022,toolllm2023,toolhop2025,taubench2025,osworld2024,swebench2024}. These benchmarks are valuable because many failures can be checked by tools, task assertions, execution results, or natural-language judges. At the same time, they expose a deployment difficulty: as tool ecosystems grow and workflows become longer, a single user request contains many heterogeneous steps, only some of which are safe for cheaper execution or specialization.

The limitations across these methods are not a lack of routers, skills, or post-training methods in isolation. The missing systems layer is a conservative mechanism that connects them inside a running agent: traces should identify repeated invocation types, train or update students on hard examples, update routers at the same granularity, and admit all changes only through replay with verifier-backed fallback. MERA addresses this gap by treating execution traces as shared supervision for SkillBook statistics, invocation-level routing, and model adaptation. Rather than only routing around a fixed small model, MERA expands cheap execution where replay shows that the combined router, skill state, and adapter preserve task quality.

%% file: method_pipeline_tikz.tex
\begin{tikzpicture}[
  x=1cm, y=1cm,
  >=Latex,
  font=\small,
  block/.style={draw=black!25, rounded corners=3pt, thick, align=center, inner sep=3.5pt, minimum height=1.05cm},
  blueblock/.style={block, fill=blue!6, draw=blue!35},
  greenblock/.style={block, fill=green!6, draw=green!35},
  orangeblock/.style={block, fill=orange!8, draw=orange!45},
  redblock/.style={block, fill=red!5, draw=red!35},
  grayblock/.style={block, fill=black!2},
  flow/.style={->, thick, draw=black!55},
  dataflow/.style={->, thick, draw=black!45}
]

\node[font=\bfseries\small, anchor=west] at (-0.15,1.4) {Runtime loop};
\node[grayblock, text width=2.05cm] (inv) at (0.85,0) {\textbf{Invocation}\\serialized prompt\\local context};
\node[blueblock, text width=2.15cm] (router) at (3.95,0) {\textbf{Router}\\strong / cheap /\\specialized student};
\node[orangeblock, text width=2.15cm] (skill) at (7.1,0) {\textbf{Skill selector}\\optional template\\dispatch};
\node[greenblock, text width=2.05cm] (exec) at (10.2,0) {\textbf{Execution}\\LLM call\\tool use};
\node[redblock, text width=1.6cm] (verify) at (12.9,0) {\textbf{Verifier}\\schema\\success\\fallback};

\draw[flow] (inv) -- (router);
\draw[flow] (router) -- (skill);
\draw[flow] (skill) -- (exec);
\draw[flow] (exec) -- (verify);
\draw[flow, bend left=22, draw=red!65!black] (verify.north west) to node[above, font=\scriptsize] {fallback} (exec.north east);

\node[font=\bfseries\small, anchor=west] at (-0.15,-2.2) {Iterative update loop};
\node[grayblock, text width=1.9cm] (trace) at (0.7,-3.85) {\textbf{Trace store}\\prompts\\outputs\\verifier logs};
\node[grayblock, text width=2.0cm] (slice) at (3.35,-3.85) {\textbf{Step slices}\\canonicalized\\invocation states};
\node[orangeblock, text width=2.0cm] (kupdate) at (6.05,-3.85) {\textbf{SkillBook}\\signature stats\\motifs};
\node[greenblock, text width=2.0cm] (sdata) at (8.75,-3.85) {\textbf{LLM update}\\hard examples\\SFT / GRPO};
\node[blueblock, text width=2.0cm] (rupdate) at (11.45,-3.85) {\textbf{Router update}\\executable labels\\threshold};
\node[redblock, text width=1.85cm] (admit) at (14.05,-3.85) {\textbf{Joint replay}\\admission /\\next round};

\draw[flow] (trace) -- (slice);
\draw[flow] (slice) -- (kupdate);
\draw[flow] (kupdate) -- (sdata);
\draw[flow] (sdata) -- (rupdate);
\draw[flow] (rupdate) -- (admit);

\node[font=\scriptsize, text=black!60] at (7.4,-4.85) {Skill$\rightarrow$LLM};
\node[font=\scriptsize, text=black!60] at (10.1,-4.85) {LLM$\rightarrow$Router};

\draw[dataflow, dashed] (admit.north) to[out=95,in=-85]
  node[right, font=\scriptsize, pos=0.35] {admit} (verify.south);

\node[grayblock, text width=13.9cm, align=left, inner sep=5pt] at (7.1,-5.85) {%
  \textbf{Iterative dependency.} Each round updates SkillBook, the LLM adapter, then the router over shared traces
  (Skill $\rightarrow$ LLM $\rightarrow$ Router). Admission depends on joint replay evaluation.};
\end{tikzpicture}

%% file: 3_method.tex
\section{Method}

\subsection{Runtime Routing}

MERA separates the serving path from the update path. At runtime, the router observes only the serialized prompt for the current invocation and selects among a strong model, a cheap model, and optionally a specialized student. A skill selector may dispatch stable templates for recurring local structure, such as formatting, extraction, or single-tool argument construction. The selected model produces an output, and a verifier checks schema validity, tool-call legality, executable tests, or downstream success. If verification fails, the invocation falls back to a stronger model and the full event is logged.

This input-only router is intentionally restrictive. It avoids coupling routing decisions to hidden agent state or implementation-specific tool traces, making the interface easier to deploy across agent harnesses. Reliability is instead protected after execution by the verifier and fallback path. The consequence is that MERA can start conservatively: early routers may prioritize low unsafe down-routing, while replay and SkillBook evidence gradually identify regions that can be served cheaply.

\subsection{Trace Products}

The update loop operates on complete traces collected from runtime execution or replay. MERA canonicalizes each trace into step slices containing the prompt, local context, tool schemas, generated output, verifier result, retry count, fallback metadata, and any skill assignment. These slices produce three evidence streams. SkillBook records success and failure statistics for recurring prompt signatures. The learned router trains on prompt text with cheap/strong labels derived from executable small-model outcomes. The LLM adapter trains on selected hard examples where cheaper execution fails or where the router and SkillBook disagree.

The same slice can therefore support different updates without forcing the components to share the same supervision format. Router examples can remain input-only, while LLM examples preserve the context needed to execute the step. Skill examples are grouped by repeated local structure rather than by label alone.

\begin{algorithm}[t]
\caption{MERA runtime and iterative update loop}
\label{alg:mera}
\begin{algorithmic}[1]
\Require router $R$, model registry $\mathcal{M}$, SkillBook $\mathcal{K}$, verifier $V$
\For{each agent invocation $x_t$}
    \State choose model $m_t \gets R(x_t)$ and optional skill $k_t \in \mathcal{K}$
    \State execute $y_t \gets m_t(x_t,k_t)$
    \If{$V(x_t,y_t)$ fails}
        \State fallback to a stronger model
    \EndIf
    \State log $(x_t,y_t,m_t,k_t,V(x_t,y_t))$
\EndFor
\For{each update round}
    \State canonicalize traces into step slices
    \State update SkillBook, router, and LLM adapter under chosen schedule
    \State run joint replay; admit only if quality is preserved
\EndFor
\end{algorithmic}
\end{algorithm}

\subsection{Scheduled Updates}

Algorithm~\ref{alg:mera} summarizes the loop. Each update round refreshes the three tracks over shared traces, and the order respects a data dependency rather than being free. The LLM adapter is trained and queried with the SkillBook procedure prepended to its prompt, so the SkillBook update must precede LLM adaptation within a cycle (Skill $\rightarrow$ LLM); training the two in parallel would fit the adapter on a stale procedure. The router is placed last so its labels reflect the current-cycle skill state and small-model outcomes. This yields the canonical Skill $\rightarrow$ LLM $\rightarrow$ Router schedule we use throughout; because the loop iterates, any residual staleness is absorbed by the next cycle.

This design lets MERA distinguish scientific effects from systems effects. Figure~\ref{fig:system_diagram} makes the dependency structure explicit: runtime execution produces trace slices, the three update tracks consume different evidence products from those slices, and replay admits only the combined state. The evaluation therefore reports direct small-model quality separately from pre-routing, verification and fallback, end-task quality, and normalized cost. This prevents a high cascade pass rate from being mistaken for either strong standalone model capability or accurate pre-routing.

\subsection{Operational Component Definitions}

In the code-generation implementation, SkillBook is external procedural prompt
memory rather than a response cache or adapter weight. The signature function
maps tasks to two coarse dataset-level keys, \texttt{humaneval} and
\texttt{mbpp}. Each rendered entry combines static task-format instructions
with bounded successful exemplars and exemplar-grounded recurring pitfalls and
patterns. It is prepended to a new task and never returns a stored answer for an
identical prompt.

Router supervision is also operationally grounded. A task is cheap-eligible
when the current small-model rollout passes the benchmark verifier and requires
escalation otherwise. The code-generation router uses frozen
Qwen3-Embedding-0.6B prompt features and a class-balanced logistic-regression
head. Repeated observations of a task are kept in the same cross-validation
group, threshold calibration uses a disjoint shard, and policy results are
reported on held-out task identifiers. This construction replaces the earlier
weak-supervision and BERT-tiny description.

Verification is benchmark-specific. For HumanEval and MBPP, generated code is
executed in an isolated Python subprocess against the benchmark-provided test
program. For TAU-2, the official evaluator checks completion using environment
state, required tool actions, and configured natural-language assertions. The
router does not call an LLM judge at inference.

\subsection{Joint Admission}

MERA admits updates through joint replay rather than isolated component metrics. SkillBook, router, and verifier evidence define easy, hard, and uncertain regions. Easy regions are candidates for cheap serving or future student admission; hard examples feed the LLM update; uncertain regions remain protected by fallback. A new router, skill state, or adapter is promoted only if replay preserves quality while reducing cost or fallback risk.

This yields a conservative deployment strategy. The runtime system can continue using strong-model fallback while update rounds search for cheaper safe regions. When an update does not improve the joint replay result, it remains an experimental artifact rather than entering the serving registry. Replay itself is only a pre-deployment gate: a production deployment still requires shadow or canary validation, drift monitoring, and rollback because an updated policy can change the traffic distribution.

%% file: 4_experiment.tex
\section{Experiments}

\subsection{Experimental Setup}

We evaluate three questions. First, does multi-cycle adaptation improve the
small model beyond a matched SFT-only control? Second, can executable
verification and fallback turn that stronger SLM into a favorable deployed
cost--quality operating point? Third, does the model-update result transfer to
a multi-turn tool-use setting under a controlled adapter-only comparison?

The code-generation study merges HumanEval and MBPP into 546 training tasks and
582 held-out evaluation tasks with disjoint task identifiers and zero exact
prompt overlap. The small model is Qwen2.5-Coder-1.5B-Instruct and the
large teacher/fallback model is GPT-5.6 Luna. We compare four-cycle SkillBook+SFT
and SkillBook+SFT+GRPO schedules using three independent training seeds. Teacher
outputs are cached so that matched runs share the same supervision. Direct-SLM
evaluation disables both routing and fallback. For deployed policies, cost is
normalized to always using the large model, with a small:large cost ratio of
$1{:}10$. Generated code is executed in an isolated Python subprocess against
the benchmark-provided test program; these are not self-generated tests.

The SkillBook uses the documented dataset-level signatures
\texttt{humaneval} and \texttt{mbpp}. Router targets are generated from
executable small-model outcomes. Frozen Qwen3-Embedding-0.6B features feed a
class-balanced logistic-regression head; repetitions are grouped during
cross-validation, and threshold calibration is disjoint from held-out policy
evaluation. Thus no result below uses the previously described UncommonRoute
weak-label set.

\subsection{Multi-Cycle Small-Model Adaptation}

\begin{table}[t]
    \centering
    \small
    \caption{Three-seed means on the 582 held-out HumanEval+MBPP tasks. Direct
    SLM pass disables routing and fallback; normalized cost is relative to
    always using GPT-5.6 Luna.}
    \label{tab:slm_adaptation}
    \begin{adjustbox}{max width=\columnwidth}
    \begin{tabular}{lrr}
        \toprule
        Policy & Pass (\%) & Cost (\%) \\
        \midrule
        Base SLM (Qwen2.5-Coder-1.5B) & 28.7 & 10.0 \\
        Multi-cycle SFT & 44.2 & 10.0 \\
        Multi-cycle SFT+GRPO & \textbf{49.7} & 10.0 \\
        Always Luna & 86.9 & 100.0 \\
        \bottomrule
    \end{tabular}
    \end{adjustbox}
\end{table}

\begin{table}[t]
    \centering
    \small
    \caption{Final-cycle routing and fallback comparison on the matched
    SFT+GRPO artifacts (three-seed mean). Thresholds are chosen on a
    calibration shard only; RouteLLM-/FrugalGPT-style rows use matched
    adaptations rather than official checkpoints.}
    \label{tab:fallback_result}
    \begin{adjustbox}{max width=\columnwidth}
    \begin{tabular}{lrr}
        \toprule
        Policy & Pass (\%) & Cost (\%) \\
        \midrule
        Always small / exact-cache+small & 49.7 & 10.0 \\
        Always Luna & 86.9 & 100.0 \\
        RouteLLM-style + fallback & 87.0 & 97.4 \\
        FrugalGPT-style response cascade & 85.5 & 106.7 \\
        MERA router + verifier fallback & \textbf{88.3} & \textbf{60.8} \\
        \bottomrule
    \end{tabular}
    \end{adjustbox}
\end{table}

\begin{table*}[ht]
    \centering
    \caption{Strict TAU-2 comparison on the fixed 35-task split. Evaluation uses
    the official environment outcome and disables SkillBook, routing, and
    fallback.}
    \label{tab:tau2_strict}
    \begin{adjustbox}{max width=0.87\textwidth}
    \begin{tabular}{llrrrr}
        \toprule
        Policy & Agent & Airline & Retail & Telecom & Overall \\
        \midrule
        Base SLM & Qwen3.5-2B & 5/9 & 7/18 & 2/8 & 14/35 (40.0\%) \\
        Larger local endpoint & Qwen3.5-4B (unadapted) & 5/9 & 9/18 & 3/8 & 17/35 (48.6\%) \\
        Trained SLM & Qwen3.5-2B + VERL GRPO & 5/9 & 10/18 & 3/8 & \textbf{18/35 (51.4\%)} \\
        \bottomrule
    \end{tabular}
    \end{adjustbox}
\end{table*}

\begin{table*}[t]
    \centering
    \caption{Finance priority-data cost planning. Break-even divides one-time training cost by estimated daily serving savings. The final column applies a conservative 70\% savings realization factor.}
    \label{tab:finance_priority}
    \begin{adjustbox}{max width=0.87\textwidth}
    \begin{tabular}{rrrrrr}
        \toprule
        Train rows & Train cost (\$) & Deploy cost (\$/hr) & Savings (\$/hr) & Break-even & 70\% break-even \\
        \midrule
        300 & 168.75 & 0.35 & 0.46 & 15.29 days & 31.96 days \\
        500 & 281.25 & 0.35 & 1.35 & \textbf{8.68 days} & \textbf{12.47 days} \\
        \bottomrule
    \end{tabular}
    \end{adjustbox}
\end{table*}

Table~\ref{tab:slm_adaptation} reports the rebuilt multi-cycle result
(final-cycle, three-seed mean). Multi-cycle fine-tuning lifts the SLM from
28.7\% to 44.2\% (SFT) and 49.7\% (SFT+GRPO). Matched GRPO exceeds SFT by
5.5--6.7 points (95\% paired-$t$ intervals exclude zero). Cascade quality is
already near 88\%; the gain is a stronger directly usable SLM with lower
fallback demand, not a cascade lift.

\subsection{Verifier-Backed Fallback}

Table~\ref{tab:fallback_result} separates direct model improvement from the
deployed operating point. Exact-cache coverage is zero on held-out prompts,
so it equals always-small. With verifier fallback, MERA matches
near-Luna quality at 60.8\% cost, while the RouteLLM- and FrugalGPT-style
baselines remain near always-Luna cost. The system win is the evolved SLM
plus verifier fallback; the learned pre-router is weak, and most quality
preservation comes from verification rather than routing alone.

\subsection{TAU-2 Adapter Ablation}

Table~\ref{tab:tau2_strict} is a clean base-versus-GRPO comparison under the
same parser, split, user simulator, and no-fallback policy. The trained
Qwen3.5-2B improves from 14/35 to 18/35 and can reach the performance of the
unadapted Qwen3.5-4B endpoint. The paired test has seven wins, three losses,
and 25 ties (one-sided McNemar $p=0.171875$), so the comparison is underpowered
and supports feasibility on tool use rather than broad cross-domain generality.

\subsection{Finance Deployment Planning}

Finance is retained as a deployment-planning case rather than a public
benchmark. Under the stated workload and serving-cost assumptions, the 500-row
adaptation setting has the larger one-time training cost but the shorter
nominal break-even time: 8.68 days, or 12.47 days when only 70\% of estimated
savings are realized. This analysis connects the paper's primary small-model
improvement objective to an operational decision about when adaptation pays
for itself.

\subsection{Discussion}

The rebuilt results place small-model evolution at the center of the paper.
The strongest evidence is the three-seed SLM lift under matched SFT and
SFT+GRPO. Verifier fallback turns that improvement into a near-Luna
cost--quality point at 60.8\% cost, while query/response routers stay near
always-Luna cost. TAU-2 supplies a clean but underpowered adapter-only check
in which a fine-tuned 2B agent matches an unadapted 4B endpoint, and finance
illustrates when adaptation pays for itself. A statistically established TAU-2
gain remains outside the evidence.

%% file: 5_conclusion.tex
\section{Conclusion}

MERA uses shared executable traces primarily to improve the small model, with
SkillBook, routing, and verifier-backed admission as supporting machinery. On
held-out HumanEval$+$MBPP, four-cycle adaptation raises Qwen2.5-Coder-1.5B from
28.7\% to 49.7\% direct pass, and verifier-backed deployment retains 88.3\% pass
at 60.8\% of always-Luna cost. On TAU-2, an adapted Qwen3.5-2B improves from
14/35 to 18/35 and matches an unadapted 4B endpoint, though the comparison is
underpowered. A finance break-even analysis further connects one-time adaptation
cost to serving savings. Overall, MERA offers a conservative protocol for raising
small-model capability from verifier-grounded traces rather than only routing
around a fixed student; broader verified workloads and multi-seed agentic
evaluation remain necessary.

%% file: appendix.tex
\section{Additional Experiment Details}

\subsection{Replay and Router Label Construction}

MERA treats replay as the common interface between routing, SkillBook updates, and model adaptation. For each logged invocation, we retain the serialized input, available tool schema, model output, verifier result, retry and fallback metadata, and any skill identifier. Replay then executes candidate runtime states against the same slice representation. A slice is considered eligible for cheaper execution only when the candidate output satisfies the same verifier used by the strong-model path. If the cheap model, adapter, or skill state fails verification, the slice remains assigned to the stronger model or the fallback region.

Router labels are therefore conservative. Positive cheap-model labels come from slices where cheaper execution preserves the verifier outcome, while strong-model labels come from failures, verifier uncertainty, or examples outside the observed support of the current SkillBook statistics. This is stricter than training a router from model confidence alone: the router is not asked to predict whether a cheap model sounds plausible, but whether the current runtime state can safely handle the invocation under replay.

\subsection{Admission Metrics}

The normalized cost metric reports the estimated serving cost of the admitted policy relative to always using the large model. Let $c_s$ and $c_l$ be the small- and large-model costs for an invocation, and let $f$ denote whether fallback is triggered. The per-invocation replay cost is counted as $c_s + f c_l$ when a cheap path is attempted and as $c_l$ when the large model is selected directly. We report the aggregate ratio over the held-out replay set. Fallback rate is the fraction of invocations for which the cheap path fails verification and requires escalation.

An update is a candidate for admission only when it improves the cost-quality operating point without introducing additional unverified failures. In the experiments, this rule is applied to the joint state rather than to isolated component metrics: a router that looks accurate in isolation is not sufficient if the resulting skill or adapter path fails replay. Conversely, an adapter improvement is reported both as a direct-SLM gain and as a deployed cascade with verifier fallback, so a weak pre-router is not mistaken for a weak student.

\subsection{Component Update Details}

SkillBook updates aggregate repeated prompt signatures and local execution motifs. The goal is not to memorize full trajectories, but to identify compact invocation types that have stable verifier outcomes across traces. The LLM update track uses hard examples from slices where the cheaper model fails or where router and SkillBook evidence disagree. The router update track consumes prompt-level labels derived from executable outcomes and can be scheduled after the other tracks when it should observe current-cycle adapter or SkillBook evidence.

Because the LLM adapter consumes the SkillBook procedure, we run the dependency-respecting Skill $\rightarrow$ LLM $\rightarrow$ Router schedule. The SFT-only and SFT+GRPO arms share the same split, cached teacher outputs, verifier, and cost accounting. Direct-SLM evaluation disables routing and fallback and is the primary scientific readout of student improvement; the routing comparison reuses the matched final-cycle artifacts as supporting deployment evidence.

\subsection{TAU-2 Controlled Protocol}

The reportable TAU-2 comparison uses the same 35 tasks for all agents, a
Qwen3.5-4B user simulator, non-thinking templates, a 1024-token cap per agent
turn, and no SkillBook, router, or fallback during evaluation. Final pass/fail
comes from the official environment outcome. The base and GRPO rows therefore
isolate the adapter under one inference protocol. The unadapted Qwen3.5-4B
endpoint is reported as a descriptive reference under the same split (17/35);
it is not part of the paired base-versus-GRPO claim. Older runs that changed the
parser, SFT recipe, user simulator, token cap, or fallback policy are excluded
rather than pooled with this comparison.

The paired base-versus-GRPO outcome contains seven wins, three losses, and 25
ties (14/35 $\to$ 18/35). We report the one-sided exact McNemar value,
$p=0.171875$, and treat the observed four-task increase as underpowered
evidence of feasibility rather than a statistically established gain.

\subsection{Finance Break-Even Calculation}

The finance table is a deployment-planning calculation rather than a public
benchmark. Nominal break-even is
\[
    \mathrm{break\mbox{-}even} =
    \frac{\mathrm{one\mbox{-}time\ training\ cost}}
         {\mathrm{daily\ serving\ savings}}.
\]
The conservative column applies a 70\% realization factor to estimated savings
before computing break-even. These assumptions keep the finance result separate
from the open-source HumanEval+MBPP and TAU-2 benchmark evidence.

\section{Limitations}

MERA is designed around replay, verification, and conservative staged deployment, and these choices introduce corresponding limitations. First, the quality of both routing labels and student admission decisions depends on verifier coverage. If the verifier fails to capture an important semantic failure mode, replay may overestimate the safety of down-routing or skill promotion. Second, our simple-step-first strategy is intentionally biased toward narrow and highly checkable slices. This makes early deployment safer, but it also means that the framework may realize its gains gradually and may leave a substantial fraction of difficult long-horizon reasoning on the strongest model for a long time.

Third, the current evaluation protocol is trace-centric rather than fully online. Replay is attractive because it enables controlled counterfactual evaluation of routing, student models, and skills under a shared verifier, but replay cannot perfectly capture distribution shift induced by changing the runtime policy itself. Fourth, skill promotion assumes that repeated local subgraphs of agent behavior can be identified and canonicalized into stable templates. Some tasks may remain too heterogeneous for this representation to pay off. Fifth, the learned pre-router remains weak in our rebuilt study: most deployed quality preservation comes from executable verification and fallback rather than accurate upfront routing. Finally, the strongest multi-seed evidence is code generation; the TAU-2 adapter check is clean but underpowered, and gains from specialization remain limited if the workload is highly non-stationary or contains few reusable step types.

\section{Broader Impact}

MERA aims to reduce the cost of agentic systems without giving up end-to-end reliability. A positive consequence is that stronger agent workflows may become deployable at lower serving cost, which could make high-quality automation more accessible. The same trace-centric design may also improve operational safety by requiring replay, verification, and fallback before new routing or specialization decisions reach production.

The same capabilities also carry risks. More efficient agents may accelerate large-scale automation in settings where reliability, privacy, or oversight matter. If verification is incomplete, down-routing or skill promotion could create subtle failures that are cheap to execute but costly to detect. The framework could also be used to lower the operating cost of agents in sensitive domains without adequate human review. These risks suggest that deployment should remain conservative, with explicit verifier coverage, staged admission, audit logs, and scope restrictions on production use.

\section{LLM Usage}

Large language models are a core methodological component of this work. They are used both as runtime policies within the agent and as the objects being routed, specialized, and evaluated. The method further relies on replay across multiple candidate models to generate routing labels and admission decisions. Our use of LLMs is therefore part of the scientific contribution itself rather than a writing-only aid.